\documentclass{article}

\PassOptionsToPackage{numbers}{natbib}
\usepackage[final]{nips_2017}

\usepackage[utf8]{inputenc} % allow utf-8 input
\usepackage[T1]{fontenc}    % use 8-bit T1 fonts
\usepackage{hyperref}       % hyperlinks
\usepackage{url}            % simple URL typesetting
\usepackage{booktabs}       % professional-quality tables
\usepackage{amsfonts}       % blackboard math symbols
\usepackage{nicefrac}       % compact symbols for 1/2, etc.
\usepackage{microtype}      % microtypography
\usepackage{comment}
\usepackage{multirow}

\title{An Encoder-Decoder Model for ICD-10 Coding of Death Certificates}

\author{
  Elena Tutubalina\thanks{This work was supported by the Russian Science Foundation grant no. 15-11-10019.}\\
  Kazan Federal University\\
  Kazan, Russia 420008 \\
  \texttt{elvtutubalina@kpfu.ru} \\
  \And
   Zulfat Miftahutdinov \\
   Kazan Federal University\\
   Kazan, Russia 420008 \\
   \texttt{zulfatmi@gmail.com} \\
}  
\begin{document}

\maketitle

\begin{abstract}
Information extraction from textual documents such as hospital records and health-related user discussions has become a topic of intense interest. The task of medical concept coding is to map a variable length text to medical concepts and corresponding classification codes in some external system or ontology. In this work, we utilize recurrent neural networks to automatically assign ICD-10 codes to fragments of death certificates written in English. We develop end-to-end neural architectures directly tailored to the task, including basic encoder-decoder architecture for statistical translation. In order to incorporate prior knowledge, we concatenate cosine similarities vector among the text and dictionary entry to the encoded state. Being applied to a standard benchmark from CLEF eHealth 2017 challenge, our model achieved F-measure of 85.01\% on a full test set with significant improvement as compared to the average score of 62.2\%  for all official participants' approaches. %We also obtained significant improvement from 26.1\% to 44.33\% on an external test set as compared to the average score of the submitted runs.
\end{abstract}

\section{Introduction}
Recent years have seen many new applications of Natural Language Processing (NLP) to biomedical information. Much of this work has been focused on a central task of information extraction, that is named entity recognition from the scientific literature or electronic health records (EHRs). %Comparatively little work has been carried out to automatically assign a formal medical concept and corresponding classification code to a free-form text fragment from EHRs. 
The task of medical concept normalization is highly important for many clinical applications in the fields of health management and patient safety.

There are several widely used ontologies of medical concepts such as the Unified Medical Language System (UMLS), SNOMED CT, and International Classification of Diseases (ICD-9, ICD-10). In particular, each medical concept in ICD is mapped onto a unique identifier which consists of a single alphabet prefix and several digits. Single alphabet prefix represents a class of common diseases (e.g. ``J'' covers diseases of the respiratory system, ``V'' covers external causes of morbidity) and digits represent specific type of disease (e.g. ``J20.2'' covers “acute bronchitis due to Streptococcus'', ``V25'' covers ``motorcycle rider injured in collision with railway train or railway vehicle''). 

In this work, we view ICD-10 coding as a sequence learning task. A sequence of codes is generated from a natural language text from medical notes by preserving the semantics of the text as much as possible. Motivated by the recent success of recurrent neural networks (RNNs), this work adopts RNN with an encoder-decoder architecture. For evaluation, we adopt a CDC corpus provided for the task of ICD-10 coding in CLEF eHealth 2017. This corpus contains free-text descriptions of causes of death in English reported by physicians. Table \ref{tab:ex} contains examples of descriptions. There are several major challenges which information extraction methods face: (i) lexical, morphological, and syntactic variants; (ii) paraphrases, synonyms; (iii) abbreviations, ambiguity; (iv) misspellings and shortened forms of words.

\begin{table}[]
\caption{Examples of raw texts from depth certificates with medical concepts and ICD codes.}
\centering
\label{tab:ex}
\begin{tabular}{|l|p{15em}ll|}
\hline
\# & \textbf{Sample} & \multicolumn{1}{r}{\textbf{Medical Concept}} & \multicolumn{1}{r|}{\textbf{Code}} \\ \hline
1 & \multicolumn{2}{l}{CKD STAGE III, CHF, SEVERE OSTEOPOROSIS} &  \\ \hline
 &  & \multicolumn{1}{r}{Chronic kidney disease, stage 3} & N183 \\
 &  & \multicolumn{1}{r}{Congestive ventricular heart failure} & I500 \\
 &  & \multicolumn{1}{r}{Osteoporosis} & M819 \\ \hline
2 & \multicolumn{2}{l}{A.FIB., D.M. TYPE II} & \\ \hline
 &  & \multicolumn{1}{r}{Atrial fibrillation} & I48 \\
 &  & \multicolumn{1}{r}{Type 2 diabetes mellitus} & E119 \\ \hline
3 & \multicolumn{2}{l}{CAD / s/p CABG / Volume overload} &  \\ \hline
 &  & \multicolumn{1}{r}{Acute coronary artery disease} & I251 \\
 &  & \multicolumn{1}{r}{Fluid overload} & E877 \\ \hline
4 & \multicolumn{2}{l}{P.V.D.} &  \\ \hline
 &  & \multicolumn{1}{r}{Peripheral vascular disease} & I739 \\
%5 & \multicolumn{2}{l}{Neutropenic fever, pneumonia} &  \\ 
%&  & \multicolumn{1}{r}{Chronic Neutropenia} & D70 \\
%&  & \multicolumn{1}{r}{Fever} & R509\\
%&  & \multicolumn{1}{r}{Pneumonia} & J189 \\  
 \hline
\end{tabular}
\end{table}

We utilize Long Short-Term Memory (LSTM) to map the input sequence into a vector representation, and then another LSTM to decode the target sequence from the vector. The network relies on two sources of information: word representations learned from unannotated corpora and a manually curated ICD-10 dictionary provided by the organizers of the task. This work is an extended version of the conference paper \cite{miftakhutdinov2017kfu}.

\section{Background}
There exist many applications where a system needs to mediate between natural language expressions and elements of a vocabulary in an ontology. Huang and Lu \cite{huang2015community} gave an overview of the work done in the organization of biomedical NLP (BioNLP) challenge evaluations up to 2014. We briefly give an overview of the major findings in previous research on terminology association.
Many BioNLP evaluations have also focused on named entity recognition (NER) of disease names in clinical notes  (e.g., ShARe/CLEF eHealth lab, SemEval 2014 lab). Automatic approaches to BioNLP tasks roughly fall into two categories: (i) linguistic approaches based on dictionaries, association measures, morphological and syntactic properties of texts \cite{van2016erasmus,mottin2016bitem,ghiasvand2014uwm,tang2014uth,cabot2017sibm}; (ii) machine learning approaches \cite{leaman2011ncbi,leaman2013dnorm,dermouche2016ecstra,zweigenbaum2016limsi,miftakhutdinov2017kfu,ebersbach2017fusion}. The CLEF Health 2016 and 2017 labs addressed the problem of mapping death certificates to ICD codes. Death certificates are standardized documents filled by physicians to report the death of a patient \cite{CLEF2017Overview}. For the CLEF eHealth 2016 lab, 5 teams participated in the shared task 2 about the ICD-10 coding of death certificates in French \cite{neveol2016clinical}. For the CLEF eHealth 2017 lab, 9 teams participated in the shared task 1 about the ICD-10 coding of death certificates in French and English \cite{CLEFTask1}. Mulligen et al. \cite{van2016erasmus} obtained the best results in task 2 by combining a Solr tagger with ICD-10 terminologies. The terminologies were derived from the task training set and a manually curated ICD-10 dictionary. They achieved F-measure of 84.8\%. Mottin et al. \cite{mottin2016bitem} applied pattern matching approach and achieved the F-measure of 55.4\%. Dermouche et al. \cite{dermouche2016ecstra} applied two machine learning methods: (i) a supervised extension of Latent Dirichlet Allocation (LDA), i.e., Labeled-LDA and (ii) Support Vector Machine (SVM) based on bag-of-words features. This study did not focus on designing effective features to obtain better classification performance. Zweigenbaum and Lavergne \cite{zweigenbaum2016hybrid} utilized a hybrid method combining simple dictionary projection and mono-label supervised classification. They trained Linear SVM on the full training corpus and the 2012 dictionary provided for CLEF participants. This hybrid method obtained an F-measure of 85.86\%. The TUC-MI team \cite{ebersbach2017fusion} utilized fusion methods in conjunction with support vector machines with a large scale feature set. The SIBM team \cite{cabot2017sibm} developed a dictionary-based approach and fuzzy matching methods. The LIMSI team \cite{zweigenbaum2017multiple} explored the combination of a dictionary-based method and SVM. Overall, most methods utilized dictionary-based semantic similarity and, to some extent, string matching.

\section{Encoder-Decoder Model}
The basic idea of our approach can be intuitively explained as follows: when we try to link a sentence to medical concepts, we do not really go word by word but rather first construct some semantic representation of this sentence and then unroll this representation in the target sequence using a neural network model. For instance, the sequence ``Neutropenic fever, pneumonia'' is mapped to ``D70 R509 J189''.  This intuition is formally captured in the encoder-decoder architecture. We adopted the architecture as described in \cite{cho2014learning}. 

RNNs are naturally used for sequence learning, where both input and output are word and label sequences, respectively. RNN has recurrent hidden states, which aim to simulate memory, i.e., the activation of a hidden state at every time step depends on the previous hidden state \cite{elman1990finding}. An important modification of the basic RNN architecture is bidirectional RNNs. One of the most widely used such modifications of RNNs is called the \emph{Long Short-Term Memory} (LSTM)~\cite{GreffSKSS15}. Our system utilizes LSTM to map the input sequence into a vector representation, and then another LSTM to decode the target sequence from the vector.  The primary goal of applying bidirectional encoder to ICD-10 coding is to capture ``semantic representation'' based on not only the past but also the future context on every time step.  We utilize left-to-right LSTM as the decoder. %We initialize word embeddings as input. 

In order to incorporate prior knowledge, we additionally concatenated cosine similarities vector between the text and dictionary's entries to the encoded state. CLEF participants were provided with a manually created dictionary. This dictionary named AmericanDictionary contains quadruplets (diagnosis text, codes Icd1, IcdC, Icd2). We presented the ICD-10 code as a single document by concatenating diagnosis texts associated with this code. In order to provide a ICD-10 code and 
an input sequence with vector representations, we computed the TF-IDF transformation and calculated the cosine similarity between these vectors. We only consider pairs (diagnosis text, Icd1) for our system since most entries in the dictionary are associated with these codes.

\section{Evaluation}
The CLEF e-Health 2017 Task 1 participants were provided with data from 13,330 and 14,833 raw texts from death certificates for training and testing, respectively. The full test set includes the ``external'' test set which is limited to textual fragments with ICD codes linked with a particular type of deaths, called ``external causes'' or violent deaths. The full set includes 18,928 codes (900 unique codes), while the ``external'' set includes only 126 codes (28 unique codes). Statistics of the corpus are presented in Table \ref{tab:corpstats}.

\begin{table}[]
\caption{Statistics of the CDC American Death Certificates Corpus from \cite{CLEFTask1}.}
\label{tab:corpstats}
\centering
\begin{tabular}{|l|r|r|}
\hline
 & \textbf{Train} & \textbf{Test} \\ \hline
Certificates & 13,330 & 6,665 \\
Lines & 32,714 & 14,834 \\
Tokens & 90,442 & 42,819 \\
Total ICD codes & 39,334 & 18,928 \\
Unique ICD codes & 1,256 & 900 \\
Unique unseen ICD codes & - & 157 \\ \hline
\end{tabular}
\end{table}

We applied the word embeddings trained on 2,5 millions of health-related reviews from \cite{miftahutdinov2017}. The embeddings were trained with the Continuous Bag of Words model with the following parameters: vector size of 200, the length of local context of 10, negative sampling of 5, vocabulary cutoff of 10. Additionally, we applied word embeddings trained on biomedical literature indexed in PubMed\cite{pyysalo2013distributional} and a part of Google News dataset\footnote{https://code.google.com/archive/p/word2vec/}. Statistics of the word embeddings are presented in Table \ref{tab:stats}. For out-of-vocabulary words with the pre-trained word model, we used representations randomly sampled.
In order to find optimal neural network configuration and word embeddings, the five-fold cross-validation procedure was applied to the training set. Embedding layers are trainable for all networks. Table \ref{tab:tabCross} shows the five-fold cross-validation results on the training dataset. It shows that all models with prior knowledge obtained better results. Models with different word embeddings obtained similar results.

\begin{table}[h]
\caption{Statistics of \textit{word2vec} embeddings.}
\label{tab:stats}
\centering
\begin{tabular}{|c|c|c|c|c|}
 \hline
\textbf{Embeddings} & \textbf{Dim.} & \textbf{\#tokens} & \textbf{\% of tokens (train data)} & \textbf{\% tokens (test data)} \\  \hline
HealthVec & 200 & 73,644 & 68\% & 70\% \\
PubmedVec & 200 & 2,351,706 & 87\% & 88\% \\
GoogleNewsVec & 300 & 3,000,000 & 73\% & 76\%  \\
\hline
\end{tabular}
\end{table}

\begin{table}[h]
\caption{Five-fold cross-validation results on the training dataset.}
\label{tab:tabCross}
\centering
\begin{tabular}{|c|c|c|c|c|}
\hline
\textbf{Encoder-decoder LSTM} & \textbf{Embeddings} & \textbf{P} & \textbf{R} & \textbf{F} \\ \hline
\multirow{3}{*}{with prior knowledge} & HealthVec & .876	&.811 &.842 \\ 
 & PubmedVec & .881&	.816& .847 \\
 & GoogleNewsVec & .879&.811& .843 \\ \hline
\multirow{3}{*}{without prior knowledge} & HealthVec & .857&	.802 & .828 \\
 & PubmedVec &.842&.796 &.819 \\ 
 & GoogleNewsVec &.844&.790 &.816 \\ \hline
\end{tabular}
\end{table}

We have implemented networks with the Keras library \cite{chollet2015keras}. We use the 600-dimensional hidden layer for the encoder RNN chain. Finally, the last hidden state of LSTM chain output concatenated with cosine similarities vector is fed into a decoding LSTM layer with 1000-dimensional hidden layer and softmax activation. In order to prevent neural networks from overfitting, we applied dropout of 0.5 \cite{srivastava2014dropout}. We used categorical cross entropy as the objective function, HealthVec as input, and the Adam optimizer \cite{kinga2015method} with the batch size of 20. We trained our model for 10 epochs.

Our neural models were evaluated on texts in English using evaluation metrics of task 1 such as precision (P), recall (R) and balanced F-measure (F). For comparison, we present our results and several official results of participants' methods (TUC-MI, SIBM teams, etc.) which did not resort to RNNs \cite{ebersbach2017fusion,cabot2017sibm,CLEFTask1} in Table \ref{tab:tab1}. Our encoder-decoder model obtained F-measure of 85.0\% on a full test set with significant improvement as compared to the average score of 62.2\% for all official CLEF participants' approaches that were based on machine learning or knowledge-based algorithms. Our model obtained comparable results with the LIMSI team that combined SVM with the dictionary for multi-label classification and submitted unofficial runs due to conflict of interest. The difference of results on two sets is explained by a small number of codes in the latter case.

% We also obtained significant improvement from 26.1\% to 44.3\% on an external test set as compared to the average score of the submitted runs. 

\begin{table}[h]
\caption{ICD-10 coding performance from \cite{CLEFTask1} on the full test set (left) and the ``external" test set (right).}
\label{tab:tab1}
\centering
\setlength{\tabcolsep}{3pt}
\begin{tabular}{|c|c|c|c|}
\hline
 & \textbf{P} & \textbf{R} & \textbf{F} \\ \hline
\multicolumn{4}{|c|}{Official runs submitted} \\ \hline
Encoder-decoder LSTM& .893  & .811 & .850\\ \hline
TUC-MI-run1 & .940 &.725& .819 \\ 
SIBM-run1 & .839 &.783 &.810\\ 
WBI-run1 & .616 & .606 & .611 \\ 
LIRMM-run1 & .691 & .514 & .589 \\\hline
Average score &  .670& .582& .622  \\ 
Median score & .646 & .606 &.611\\ \hline
\multicolumn{4}{|c|}{Non-off} \\ \hline
LIMSI & .899 & .801 & .847 \\ \hline
\end{tabular}
%\caption{ICD-10 coding performance on the ``external" test set.}
\begin{tabular}{|c|c|c|c|}
\hline
 & \textbf{P} & \textbf{R} & \textbf{F} \\ \hline
\multicolumn{4}{|c|}{Official runs submitted} \\ \hline
Encoder-decoder LSTM & .584 & .357 & .443 \\ \hline
TUC-MI-run1 & .880 &.175 &.291 \\ 
SIBM-run1 & .426 &.389& .407\\ 
WBI-run1 & .246& .119 &.160 \\ 
LIRMM-run1 & .232& .524& .322 \\\hline
Average score & .405& .267& .261 \\ 
Median score &.279 &.262& .274 \\ \hline
\multicolumn{4}{|c|}{Non-off} \\ \hline
LIMSI & .723& .373 &.492 \\ \hline
\end{tabular}
\end{table}

\section{Conclusion}
In this work, we have applied deep neural networks, in particular, LSTM-based encoder-decoder architecture, to the problem of ICD-10 coding. We
have obtained very promising results, both quantitatively and qualitatively. We outline three directions for future work. First, the use of novel architectures and multilingual neural networks remains to be explored. We would like to explore alternative distributed word representations trained on medical notes from electronic health records. Second, a promising research direction is the integration of linguistic knowledge into the models. Third, future research might focus on developing extrinsic test sets for medical concept normalization.

\bibliographystyle{unsrt}
%\bibliography{ml}

\end{document}